# Role of Deep LSTM Neural Networks And Wi-Fi Networks in Support of Occupancy Prediction in Smart Buildings

Basheer Qolomany[1], Ala Al-Fuqaha[1], Driss Benhaddou[2], Ajay Gupta[1]

[1]Computer Science Dept., College of Engineering and Applied Sciences, Western Michigan University, Kalamazoo, Michigan, USA
{basheer.qolomany, ala.al-fuqaha, ajay.gupta}@wmich.edu

[2]Engineering Technology Dept., College of Technology, University of Houston, Houston, Texas, USA
dbenhadd@central.uh.edu

*Abstract*— Knowing how many people occupy a building, and where they are located, is a key component of smart building services. Commercial, industrial and residential buildings often incorporate systems used to determine occupancy. However, relatively simple sensor technology and control algorithms limit the effectiveness of smart building services. In this paper we propose to replace sensor technology with time series models that can predict the number of occupants at a given location and time.

We use Wi-Fi datasets readily available in abundance for smart building services and train Auto Regression Integrating Moving Average (ARIMA) models and Long Short-Term Memory (LSTM) time series models. As a use case scenario of smart building services, these models allow forecasting of the number of people at a given time and location in 15, 30 and 60 minutes time intervals at building as well as Access Point (AP) level. For LSTM, we build our models in two ways: a separate model for every time scale, and a combined model for the three time scales. Our experiments show that LSTM combined model reduced the computational resources with respect to the number of neurons by 74.48 % for the AP level, and by 67.13 % for the building level. Further, the root mean square error (RMSE) was reduced by 88.2% - 93.4% for LSTM in comparison to ARIMA for the building levels models and by 80.9 % - 87% for the AP level models.

*Keywords - Time series, Machine Learning, ARIMA, LSTM, Smart Buildings, Smart Homes, IoT services, Wi-Fi networks.*

## I. INTRODUCTION

Being able to accurately count the number of occupants in a smart building has high utility for a number of applications. Information on building occupancy can be used to save energy by controlling temperature and ventilation more accurately. Number of users in the environment is important to accurately recognize the activities of (groups of) agents [1]. In the event of an emergency, first responders often need to know if people are trapped and where they might be located in large buildings. In the context of large facilities like conference centers or in the retail space, knowing how many people are in certain locations and how long they dwell can be used to value shelf-space or storefront locations [2].

Commercial, industrial and residential buildings often incorporate many approaches to determine occupancy including: passive infra-red (PIR) sensors, ultrasonic ranging sensors, microwave sensors, smart cameras, break beam sensors and laser range-finders [3]. However, these sensors extend across a wide range of cost and performance. The ability to determine the actual number of occupants in a place is often beyond the range of current common sensing techniques. Low-cost sensors, like PIR and ultrasonic ranging sensors are typically error-prone and usually only detect binary occupancy values rather than estimating load. Expensive sensors tend to require the complicated site-specific installation and standardization methods [2]. Motion detectors have inherent limitations when occupants remain relatively still [4]. Furthermore, distant passersby and wafts of warm or cold air are interpreted as motion leading to false positives [5]. Video cameras raise privacy concerns and require large amounts of data storage and complex video processing [6]. Other work has focused on the use of carbon dioxide ($CO_2$) sensors in conjunction with building models for estimating the number of people generating the measured $CO_2$ level [7]. Smart buildings have a high degree, if not full, Wi-Fi coverage and thus this paper explores the use of Wi-Fi as sensory data to predict occupancy using Wi-Fi beacon time series. In particular, we use the data to predict the number of occupants in building using Long-Short Term Memory (LSTM) deep learning networks.

Recently, time series methods have been successfully applied in a wide range of IoT application that have a time dimension such as energy usage prediction, non-intrusive activity detection, demand side management and control [8]. One of the main purposes of time series data is that past observations of the data can be used to forecast future values. The use of observations from a time series available at time $t$ to predict its value at time $t + l$ is called forecasting. In this paper we develop and compare two time series models, Auto Regression Integrating Moving Average (ARIMA) model and Long Short-Term Memory (LSTM) deep recurrent neural network to forecast the number of occupants in a smart building at a specific time under three time scales, namely, 15, 30 and 60 minutes. We focus on the following three questions:
- How to predict the number of occupants using Wi-Fi beacons?
- Which model yields better accuracy, ARIMA or LSTM?

- For LSTM models, which is better -- to build a separate model for a specific time scale or to build a combined model for the three time scales with respect to computational performance and accuracy?
- For ARIMA and LSTM models, which way is better -- to build one model for the entire building for a specific time scale or to build one model for every AP (Access Point) in the building for a specific time scale?

This is the first paper exploring the use of Wi-Fi beacons to predict the number of occupants in a commercial building, to the best of our knowledge. The advantage of using Wi-Fi is that there is no need to deploy an infrastructure to count the number of occupants in a building which reduces the cost and the complexity of the system. Another contribution of this paper is, answering the above questions which are crucial for improving the state-of-the-art in time series forecasting.

The remainder of this paper is organized as follows: Section II presents the most recent related work. In Section III, we discuss ARIMA and LSTM models. Section IV presents our experimental results and the lessons learned, and finally, Section V concludes the paper.

## II. RELATED WORK

A significant amount of work has been done in the past two decades to enable accurate and robust occupants counting. In brief, Krumm and Brush [9] presented an occupancy prediction algorithm that gives probabilities of occupancy at different times of day. However, this algorithm computes a representative Sunday, Monday, etc. for each day of the week, without being able to respond to changing occupancy patterns as PreHeat [11] does. Lu et al. [10] formulate a hidden Markov model (HMM) to predict occupancy and control HVAC systems. They collected occupancy data in eight US households for one to two weeks. Using leave-one-out cross-validation to train and test the HMM, they evaluate their approach's MissTime (i.e., total occupied time not at set point) and energy savings for each day in a week using the US Dept. of Energy's EnergyPlus simulator. Mozer et al. [11] describe a "Neurothermostat" which utilizes a hybrid occupancy predictor, making use of an available daily schedule and a neural network which was trained on five consecutive months of occupancy data. Mozer et al. show that the Neurothermostat results in a lower unified cost, where energy and occupant "comfort" are expressed. Recently, some studies focused on counting pedestrians with binary sensors and Monte-Carlo methods [12] but those are once again hardly usable in homes as they make use of an important number of landmarks such as doors, stairs and elevators, that may not be present in regular homes. Yang et al. [13] propose image-based counting technique that uses a network of simple image sensors. They introduce a geometric algorithm that computes bounds on the number and possible locations of people using silhouettes computed by each sensor through background subtraction.

In this paper we use ARIMA and LSTM time series algorithms to predict the number of occupants in a smart building using Wi-Fi network data. To the best of our knowledge, this work is the first attempt that addresses the role of time series methods to forecast the number of occupants in the smart building. One of the main advantages of this technique is that it does not require an additional infrastructure to be able to count the number of people in a building.

## III. MODELS

Time series data is any data that has a timestamp, such as IoT device data, stocks, and commodity prices. Different time series prediction models exist that work on different patterns. In this paper our focus is on an autoregressive integrated moving average (ARIMA) model as this is one of the most widely used statistical models for time series forecasting and thus has a strong potential for occupancy prediction using Wi-Fi time-series data. Deep recurrent neural networks have recently gained a lot of attention in exhibiting good prediction capabilities. Therefore, we also focus on the Long Short-Term Memory (LSTM) model of deep recurrent neural networks.

**3.1 ARIMA time series model**

ARIMA is one of the most common univariate time series models, it is also well-known as Box-Jenkins methodology in the model selection procedure, and the popularity of this model is due to its statistical properties [14]. The AutoRegressive (AR) part of ARIMA indicates that the variable of interest is regressed on its lagged values. The Moving Averages (MA) part indicates that the regression error is actually a linear compound of error terms whose values occurred simultaneously and at various times in the past. The Integrated (I) indicates that the data values have been replaced with the difference between their values and the previous values [15] [16].

An AutoRegressive of order p, AR(p), component of an ARIMA model is a discrete time linear equations with noise. It can be written in the form:

$$X_t = \sum_{i=1}^{p} \alpha_i X_{i-1} + Z_t \qquad (1)$$

Where the terms in α are autocorrelation coefficients at lags 1,2...p and $z_t$ is a residual error term. Note that this error term specifically relates to the current time period *t*.

A Moving Average with order q, MA(q), model can be used to provide a good fit to some datasets. A simple form of such models, based on prior data, can be written as:

$$X_t = \sum_{i=0}^{q} \beta_i Z_{i-1} \qquad (2)$$

Where the $\beta_i$ terms are the weights applied to prior values in the time series, and it is usual to define $\beta_i=1$, without loss in generality.

An AutoRegressive Moving Average with orders p and q, ARMA(p, q), is the one where these two models are combined by simply adding them together as a model of order (*p*,*q*), where we have *p* AR terms and *q* MA terms. An ARMA discrete time linear equation with noise has the following form:

$$X_t = \alpha_0 + \sum_{i=1}^{p} \alpha_i X_{i-1} + \sum_{i=0}^{q} \beta_i Z_{t-i} \quad (3)$$

where X, is a stationary stochastic process with non-zero mean, $\alpha_0$ is constant term, and $Z_t$, is a white noise disturbance term. A time series is said to be stationary, if the mean of the series and the covariance among its observations do not change over time and do not follow any trend [17]. The ARIMA model which generally overcomes the limitation of non-stationary time series by introducing a differencing process of subtracting the observation in the current period from the previous one. This process effectively transforms the non-stationary data into a stationary one [18]. Hence, the ARIMA model is called "Integrated" ARMA because of the stationary model that is fitted to the differenced data that has to be summed or integrated in order to provide a model for the original non stationary data. Eq. 3 denotes by notation ARIMA *(p,d,q)* where p is the order of the autoregressive part, which is the number of dependent variable lagged in the right hand side, *d* is the order of differencing performed on X, before estimating the above model, and *q* is the order of the moving-average process, which is the lagged error term in the right hand side of Eq. 3. The AR part of the model indicates that the future values of $X_t$ are weighted averages of current and past realizations. Similarly, the MA part of the model shows how current and past random shocks will affect the future values of $X_t$. The more general ARIMA process model can be written as an AR if the MA process is invertible. One of the best ways to make a series stationary on variance is through transforming the original series through log transform.

**3.2 Deep LSTM Model**

The Long Short-Term Memory network, or LSTM network, is a special kind of recurrent neural network (RNN) developed in 1997 by Hochreiter & Schmidhuber [19]. LSTM is trained using Back propagation Through Time (BPTT) and overcomes the vanishing and exploding gradient problem in standard RNN by learning tasks involving long term dependencies. Instead of neurons, LSTM networks have memory blocks that are comprised of memory cell units that are able to remember the value of a state for an arbitrary long time, as well as three different gate units that can learn to keep, utilize, or destroy a state when appropriate. The memory blocks are connected through layers [20]. We can make a deep LSTM by stacking multiple LSTM layers. Although LSTM networks are already deep architectures in the sense that they can be considered as a feed-forward neural network unrolled in time where each layer shares the same model parameters. But the deep LSTM has an additional meaning; it has been argued that deep layers in LSTM allow the network to learn at different time scales over the input. With the deep LSTM, the input to the network at a given time step goes through multiple LSTM layers in addition to propagation through time and LSTM layers [21]. Figure 1 shows the architecture for one module in an LSTM network. The network takes three inputs. $X_t$ is the input of the current time step. $h_{t-1}$ is the output from the previous LSTM unit and $C_{t-1}$ is the "memory" of the previous unit. As for outputs, $h_t$ is the output of the current network. $C_t$ is the memory of the current unit. While the internals of the module are as follows:

$$f_t = \sigma(W_{xf}x_t + W_{hf}h_{t-1} + W_{cf}c_{t-1} + b_f) \quad (4)$$
$$i_t = \sigma(W_{xi}x_t + W_{hi}h_{t-1} + W_{ci}c_{t-1} + b_i) \quad (5)$$
$$c_t = f_t c_{t-1} + i_t tanh(W_{xc}x_t + W_{hc}h_{t-1} + b_c) \quad (6)$$
$$o_t = \sigma(W_{xo}x_t + W_{ho}h_{t-1} + W_{co}c_t + b_o) \quad (7)$$
$$h_t = o_t tanh(c_t) \quad (8)$$

$\sigma$ is the logistic sigmoid function, and *f, i, c and o* are respectively the forget gate, input gate, cell state and output gate. $W_{cf}$, $W_{ci}$, and $W_{co}$ are denoted weight matrices for peephole connections. $b_f$, $b_i$, $b_c$, and $b_o$ are respectively the bias vectors for forget, input, cell state and out gates.

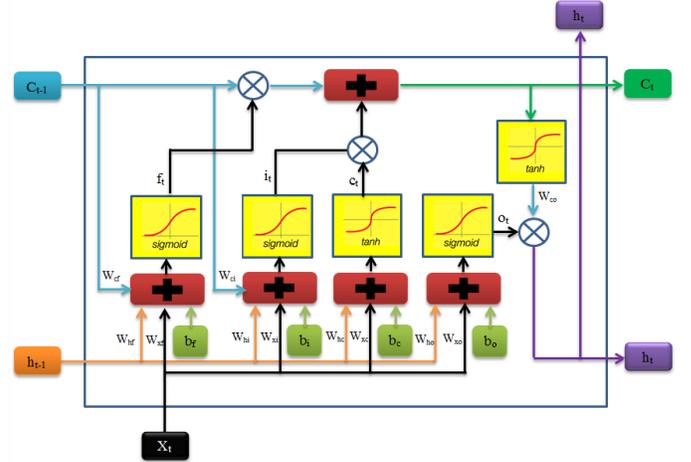

Fig. 1: LSTM Module.

In LSTM, three gates (*f, i, o*) control the information flow. The forget gate determines whether to pass the previous memory $h_{t-1}$. The ratio of the previous memory is calculated in equation (4) and used for equation (6). The input gate decides the ratio of input. When calculating the cell state, this ratio has an effect on equation (6). The output gate determines whether to pass the output of memory cell or not [22]. Equation (8) captures this process.

While building an LSTM model, one of the difficulties is tuning the numerous parameters when it comes to train the model because there is no good theory on how to do it**.** One way to tune the parameters is by applying grid search. Or use a systematic method to explore different configurations for the network. Qolomany et al. [23] proposed a systematic way to tune the deep neural network parameters using particle swarm optimization. Since our goal in this paper is to asses efficacy of LSTM and compare it to ARIMA, we simply use grid search to tune the number of hidden layers, number of neurons in a layer, batch size, number of epochs, and lag size parameters that play the major role in building a time series model using LSTM algorithm. In the experimental results section we discuss more on the role of each parameter to decrease error while building the model.

The computational complexity in terms of the total number of neurons needed to build an LSTM model for a time scale is

$$T_s = N.H + I + 1 \qquad (9)$$

Where $T_s$ is the total number of neurons needed to build a separate model with single output, $N$ is the number of neurons in each layer, $H$ is the number of hidden layers, and $I$ is the number of inputs features represented here as the lag size. While the number of neurons needed to build a single combined model for the three time scales can be calculated as

$$T_c = N.H + m.I + m \qquad (10)$$

Where $T_c$ is the total number of neurons needed to build a single combined model with multiple output, $N$ is the number of neurons in each layer, $H$ is the number of hidden layers, m is the number of different models, which is three in our case, each model represents a specific time scale (15, 30 and 60 minutes time scales), and $I$ is the number of inputs features represented here as the lag size.

### 3.3 Forecast evaluation methods

The criterion that we use to make comparisons between the forecasting ability of the ARIMA time series models and LSTM models is the root mean square error (RMSE), a standard statistical metric to measure model performance. RMSE represents the sample standard deviation of the differences between predicted values and observed values. These individual differences are called residuals when the calculations are performed over the data sample that was used for estimation [24]. The formula for RMSE is

$$RMSE = \sqrt[2]{\frac{1}{N}\sum(P_t - A_t)^2} \qquad (11)$$

Where $P_t$ is the predicted value for time t, $A_t$, is the actual value at time t, and $N$ is the number of predictions.

## IV. EXPERIMENTAL SETUP

### 4.1 Dataset

The data set is collected from the Wi-Fi network at the University of Houston main campus. The campus has full Wi-Fi coverage with Access Points (AP) managed through controllers that manage hand off allowing users to move around without losing the connection. Whenever a user carrying a Wi-Fi enabled device is passing by the network, the device is automatically exchanging beacons with the Wi-Fi network regardless whether the user is actively using the device. The Wi-Fi network captures these beacons and archive them in a storage device for further analysis. The data collected includes, connection time, connection duration, MAC address, access point ID.

In our experiments we used 6 weeks (January 15, 2016 – Feb 29, 2016) of Wi-Fi dataset for a building that has 18 AP from the University of Houston campus. We preprocessed the raw dataset as time series format for a use case scenario of services for smart buildings environment. We predict the number of occupants at a given time and location. All our experiments are conducted using ARIMA time series packages in R and Keras package under Theano platform in Python. We run our experiment on a server that has 24 cores of 2.40GHz Intel(R) Xeon(R) CPU and 32 GB memory.

### 4.2 Data pre-processing and preparation

The LSTM and ARIMA algorithms like most of other machine learning algorithms require preparing and preprocessing the raw data into a specific form in order to get the best results. First, we prepared the raw Wi-Fi dataset as a time series format for every time scale (e.g. 60, 30 and 15 minutes), such that the dataset has time with the corresponding number of occupants at that period of time. Then before feeding the data to the model, we transform the time series format into a supervised learning format by dividing it into input and output components. For time series we can achieve this by using the observation from the last time step (t-1) as the input and the observation at the current time step (t) as the output. Because the Wi-Fi dataset that we have is not stationary, the next step of preparing the dataset is to make it stationary by removing trends in the non-stationary data. We transform the time series data into stationary time series data by differencing the data. That is the observation from the previous time step (t-1) is subtracted from the current observation (t). The last step for preparing the time series data is to scale the data. We make the scaling for LSTM different than ARIMA models. The LSTM models like other neural networks expect data to be within the scale of the activation function used by the network. The default activation function for LSTM is the hyperbolic tangent (*tanh*), which outputs values between -1 and 1. So we scaled the time series data for LSTM into the range -1 and 1. While in the case of ARIMA models we use common logs transform for scaling, due to the popularity of log-returns it is easy to aggregate the log-returns over time.

### 4.3 Experimental Results

In our experiments we build and compare two time series models ARIMA and LSTM to forecast the number of occupants in the smart building at a specific time using three time scales, namely 15, 30 and 60 minutes duration. We build models for the whole building as well as for an individual AP level. In the case of LSTM, we further build our models in two ways, building a separate model for every time scale and building a single combined model for all the three time scales. Combined model is built as many to many architecture of LSTM, such that we feed the network with the inputs of every time scale but all the three models share the same hidden layers, and the outputs of this combined model are three outputs, each representing the output of a specific time scale model. As a clarification for the shortcomings of the labels in Figures 2-7, Com: combined model, Sep: separate model, AP: the models that are on the Access Point level, and Bld: the models that are on the Building-level. For example, LSTMCom15Bld refers to the LSTM combined model for the 15 time scale duration on the Building-level; LSTMSep30AP refers to the LSTM separate model for the 30 time scale duration on the AP-level; and ARIMA60Bld refers to the ARIMA model for the 30 time scale duration on the AP-le Building-level.

Figures 2 compares separate models and combined models with respect to the reduction of the root mean square error (RMSE) and computationally in terms of the number of neurons used to build such models for the three time scales on the building level and on the AP level.

We use grid search to explore the best LSTM model configurations for the number of hidden layers, number of neurons in a layer, batch size, number of epochs, and lag size parameters that play a major role in building a time series model for the LSTM algorithm. Table I shows the best configurations to build the building-level as well as the AP-level models for the three time scales in case of LSTM combined and separate individual time scales. So according to Eq. 9 the total number of neurons needed for the three separate individual time scale models for building level prediction model is $(3* 48 + 24 + 1) + (3* 32 + 48 + 1) + (2* 48 + 12 + 1) = 423$ neurons. And the number of neurons needed for the three separate individual time scale models in case of AP level prediction models is $(3* 16 + 48 + 1) + (2* 48 + 24 + 1) + (4* 48 + 24 + 1) = 435$ neurons. While according to Eq. 10 the total number of neurons needed to build a single combined model for the three time scales in case of building level prediction is $2*32 + 3*24 + 3 = 139$. And the total number of neurons needed to build a single combined model for the three time scales in case of AP level prediction is $3*32 + 3*4 + 3 = 111$.

As it is shown in Figure 2 (b), by building a single combined model for the three time scales we reduced the number of neurons by 74.48 % in case of AP level, and by 67.13 % in case of building level occupancy prediction. And at the same time as Figure 2 (a) shows, by building a single combined model we reduced the RMSE by 17.14 % - 41.33 % in case of building level models and by 20.64 % - 40.15 % in case of AP level models, except an anomaly for the AP-level 60 minute time scale model where the RMSE is increased by 16 %. Therefore, without loss in generality, we can observe that the LSTM combined model performs better – at least from a computation resource point of view. Next, we compare combined LSTM model with the ARIMA (there is no need to compare LSTM individual time scale models because LSTM combined seems to perform better).

Figure 3 shows the comparison of LSTM models with the corresponding ARIMA models for the three time scales for both the building-level as well as the AP-level prediction models. LSTM models exhibit RMSE reduction by 88.2 % - 93.4 % in case of building level models, and by 80.9 % - 87 % in case of AP level models when compared to ARIMA models. LSTM seems to outperform ARIMA.

Figures 5-7 show the effect of the parameters (number of lags, number of neurons in a layer, and number of hidden layers) on reducing the RMSE in case of LSTM models in (a) building level models and (b) AP level models. In every case we fix two parameters and set their values (to the ones that we found are the best in terms of reducing the RMSE using grid search) and start changing the third parameter. For instance, in Figure 5, we are fixing the number of neurons in a layer, and the number of hidden layers to the best corresponding model values that is listed in the Table I, and start changing the number of lags. It can be easily seen from Figures 5-7 that the RMSE values in case of building level models are always greater than the RMSE values for the corresponding time scale models for the AP-level prediction (as expected because of the fine grain modeling). However, as Figure 4 shows, there is a significant computational saving (of almost 94.28%) in terms of the number of neurons needed to build the models for building-level prediction instead of the AP level.

**4.3 Lessons learned**

We can conclude the following based on the results presented in this paper:

- Wi-Fi is a practical way to count the number of occupants in building. This information can be utilized for emergency management as well as energy efficient applications. The main advantage of this application is that it does not require an additional infrastructure to count the number of people. One of the challenges is that those who are not carrying a device will not be counted. This could be another research project that estimate the number of occupants that are missed and make it part of the overall accuracy calculation.

- As Figures 2 and 3 show LSTM models beat ARIMA models in all three time scale models. LSTM models reduced the RMSE by 88.2 % to 93.4 % on the building level models. And by 80.9 % to 87 % on the AP level when compared to ARIMA models.

- Comparing LSTM separate individual time scale models with the combined model, we find that training a single combined model for different time scales is better in terms of achieving less RMSE and better use of computation resources.

- In case of LSTM and ARIMA models, the decision to build a model on the building-level vs. AP-level is application dependent. The applications that care more about localization of the occupants in the building, it is better to build the models on the AP-level, while the applications that care more about the computational resources and energy savings and less about the accuracy,

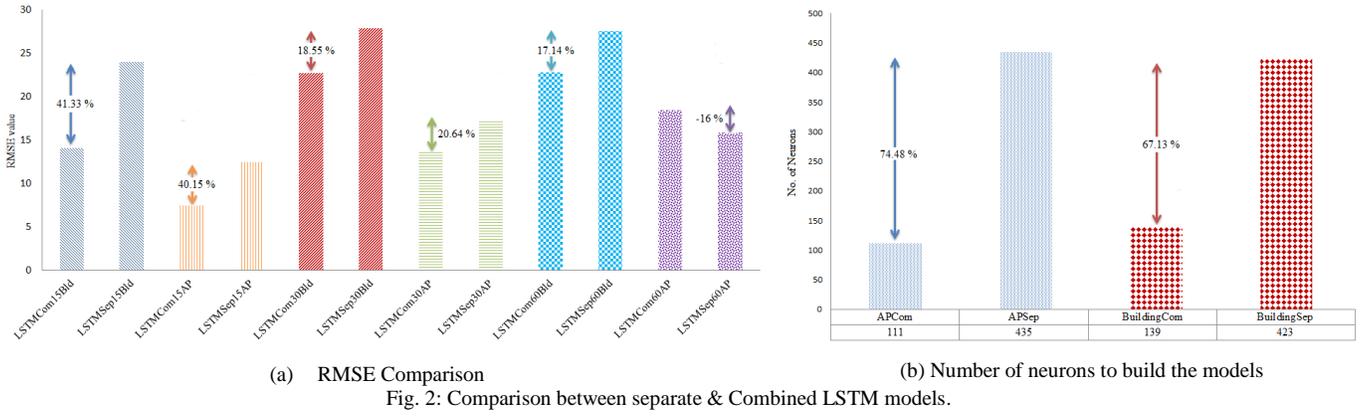

(a) RMSE Comparison

(b) Number of neurons to build the models

Fig. 2: Comparison between separate & Combined LSTM models.

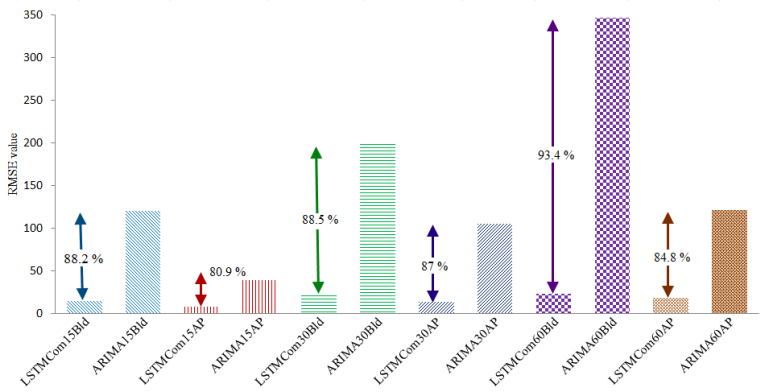

Fig. 3: Comparison between LSTM & ARIMA models.

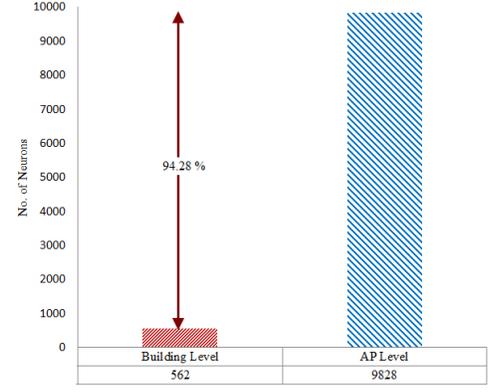

Fig. 4: Total number of neurons used to build the three time scales models for the building level vs. the AP level.

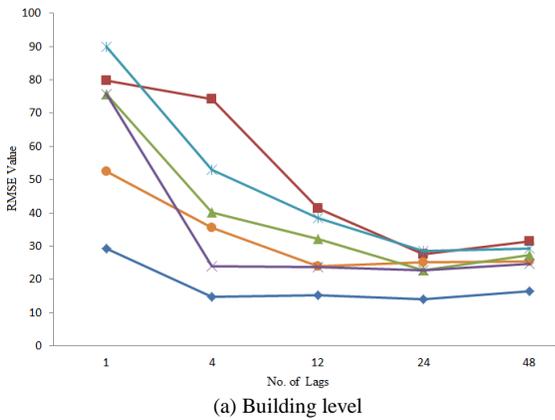

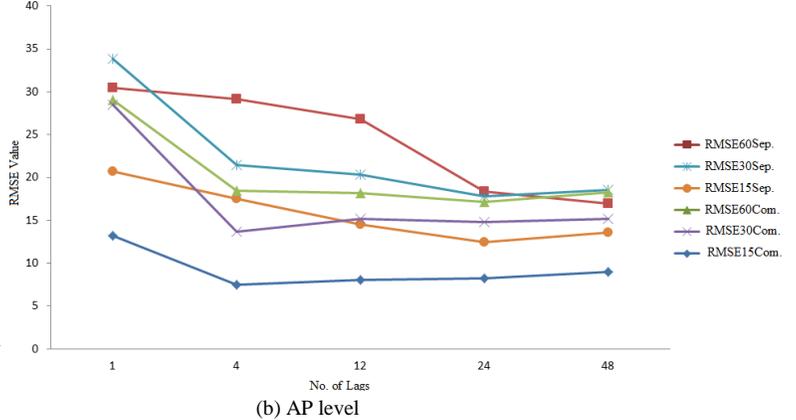

(a) Building level

(b) AP level

Fig. 5: The effect of the lag size on reducing the RMSE in LSTM models.

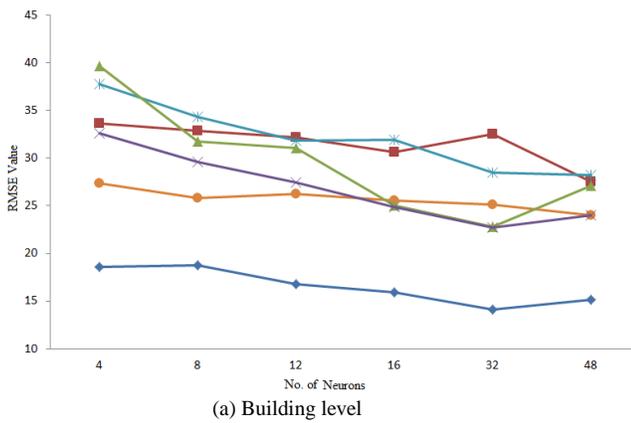

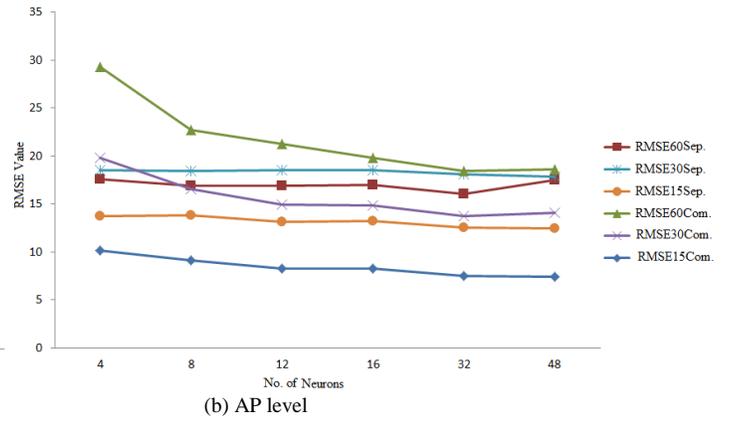

(a) Building level

(b) AP level

Fig. 6: The effect of the number of neurons on reducing the RMSE in LSTM models.

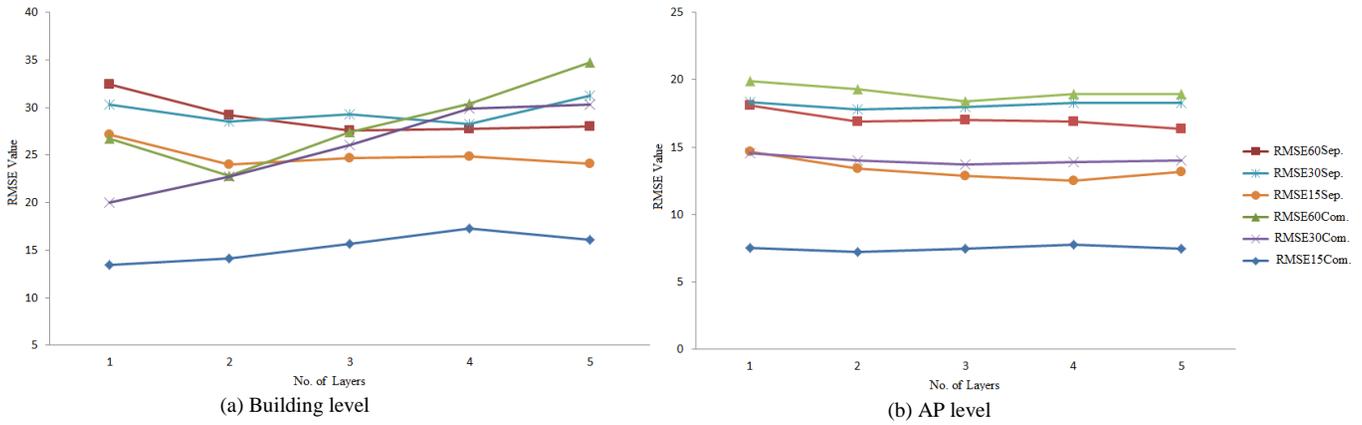

Fig. 7: The effect of the number of hidden layers on reducing the RMSE in LSTM models.

TABLE I : THE BEST CONFIGURATIONS FOR LSTM COMBINED AND SEPARATE MODELS FOR THE THREE TIME SCALES 60, 30, AND 15 MINUTES FOR THE WHOLE BUILDING AS WELL AS THE AP LEVEL. (COMBBUILDING REFERS TO BUILDING-LEVEL MODEL FOR THE LSTM COMBINED; SEP30AP REFERS TO THE ACCESS-POINT MODEL FOR THE 30 TIME SCALE DURATION, ETC.)

|  | CombBuilding | Sep60Building | Sep30Building | Sep15Building | CombAP | Sep60AP | Sep30AP | Sep15AP |
|---|---|---|---|---|---|---|---|---|
| Neurons | 32 | 48 | 32 | 48 | 32 | 16 | 48 | 48 |
| Layers | 2 | 3 | 3 | 2 | 3 | 3 | 2 | 4 |
| Lags | 24 | 24 | 48 | 12 | 4 | 48 | 24 | 24 |
| Batch size | 16 | 16 | 16 | 16 | 16 | 16 | 16 | 16 |
| Epochs | 1000 | 1000 | 1000 | 1000 | 1000 | 1000 | 1000 | 1000 |

it is better to build the models on the building level.

## V. CONCLUSIONS

This study built and compared two of state-of-the art time series prediction methods in statistics and machine learning, namely LSTM and ARIMA models based on Wi-Fi networks as an infrastructure to forecast the number of occupants at a given time and location in smart building environments. Results showed that with LSTM combined strategy we are able to reduce the number of neurons needed to build the models for every time scale by 67.13% - 74.48% when compared to individual time scale models. LSTM forecasts were considerably more accurate than those of the traditional ARIMA models, our observations revealed a RMSE reduction of 80.9% - 93.4% by LSTM. Although LSTMs are able to achieve a lower RMSE, they are extremely slow to train, can take a long time to run, often require more data to train than ARIMA models, and have lots of input parameters to tune.